# Towards Ontologically Grounded and Language-Agnostic Knowledge Graphs

Walid S. Saba

Institute for Experiential AI
Northeastern University
`w.saba@northeastern.edu`

**Abstract**

Knowledge graphs (KGs) have become the standard technology for the representation of factual information in applications such as recommendation engines, search, and question-answering systems. However, the continual updating of KGs, as well as the integration of KGs from different domains and KGs in different languages, remains to be a major challenge. What we suggest here is that by a reification of abstract objects and by acknowledging the ontological distinction between concepts and types, we arrive at an ontologically grounded and language-agnostic representation that can alleviate the difficulties in KG integration.

## 1 Introduction

Knowledge graphs are by now the standard representation of knowledge repositories that are used in various applications, such as search, recommendation engines, and question-answering systems. While there are powerful KG tools, the semantic and conceptual side of KG technology is still partially ad-hoc. In particular, the continuous update and KG integration remain to be a challenge.

A Knowledge graph (KG) is a graph structure that can be viewed as a set of triples $\langle e_1, r, e_2 \rangle$ relating real-world entities $e_1$ and $e_2$ by a relation $r$ to represent a real-world fact, as in the following examples:

(1) ⟨*RogerWaters*, *BornOn*, *01/08/1955*⟩
(2) ⟨*PinkFloyd*, *StartedIn*, *London*⟩
(3) ⟨*BarakObama*, *LivesIn*, *WhiteHouse*⟩

From the triples above that we might have in some knowledge graph $KG_1$ we can immediately point to several issues that pose major challenges in constructing and maintaining KGs. We discuss these issues next.

## 2 Alignment and Continuous Change

Here are the main issues in the triples (1) through (3) above: First, in another knowledge graph $KG_2$ that we might want to integrate with $KG_1$ there might be another *Roger Waters* where the two entities might or might not be the same and thus an entity alignment must occur with the triple in (1). Another issue here is that the triple in (2) uses "*StartedIn*" to represent the fact that the Pink Floyd band started in London. Another KG might, instead, use the relation "*FormedIn*" and a match and an alignment between the two relations is needed. Finally, the integration of

KG₁ with another KG might reveal that the triple in (3) is no longer valid and must thus be fused with new and updated information. At a minimum, then, the process of fusing together two or more KGs will first of all involve a tedious process of entity alignment (EA) (Zhang et. al., 2022), but more generally it will involve a process of continuous updating of information (Wang, et. al., 2022). Note that updating information and entity alignment both involve identifying if entities are the same (or not), where in one case we will perform a 'merge' and in the second an update.

Clearly then entity alignment is the most basic operation in any KG integration, and as such it has received the most attention. To match an entity $e_1$ in KG₁ with an entity $e_2$ in KG₂ embeddings in low dimensional space for both entities are constructed using neighboring information: related entities, immediate relations, and attributes. Entities $e_1$ and $e_2$ are considered to have a match if their vector similarity is above a certain threshold. As such, different alignment techniques mainly differ in how the embeddings are constructed. In particular, they differ in what information is bundled in the embedding, and how far in the graph are other entities, relations and attributes are still considered to be in the "neighborhood". Zhu et. al. (2021), for example, report that spreading entity information across all relations, gathering information, and bringing it back to an entity's embedding, improves embedding similarity and entity alignment. In (Lin, Y. and Liu, Z. et. al., 2016) it is further suggested that including all attributes and their values will also improve an entity's embedding. Other approaches (e.g., Zhu et. al., 2023) will also include, besides attribute values, all string information corresponding to entity, relation, and attribute names. In all these approaches the ultimate goal is to improve the construction of entity embeddings, in the hope of improving the accuracy of entity alignment (i.e., entity matching). See (Zhang, R. et. al., 2022) for a good survey of various alignment techniques.

## 3  Reifying Abstract Objects

Regardless of the novelty and the progress made by various entity alignment algorithms, the accuracy of merging different knowledge graphs, especially ones that are continuously updated, will remain to be less than desired. In this section we will argue that the problem is to be handled not with constructing ever more reliable embeddings leading to more accurate alignments, but with how knowledge graphs are constructed in the first place. Specifically, we suggest that the answer lies in proposals that have been made in the study of semantics and formal ontology. In particular, we will appeal to conceptualism and the conceptual realism of Cocchiarella (2001), where we reify (or 'object-ify') abstract concepts in a manner that is consistent with our basic "cognitive capacities that underlie our use of language". This is essentially an extension of Davidsonian semantics (Davidson, 1967; Larson, 1998) where events are treated as entities, and is also in line with Moltmann's (2013) arguments that the ontology of natural language admits references to "tropes", which are particular instances of properties.

Let us make all of this clear with an example. Consider the knowledge graphs in figure 1 where we are representing the facts expressed by "*The musician Roger Waters was born in Great Bookham on 01/08/1955*". The knowledge graph in figure 1b has the same facts expressed in figure 1a but in an ontologically grounded and linguistically agnostic representation. First, note that instead of the ad-hoc naming of relations in 1a (e.g **bornIn** and **bornOn**), in 1b we have primitive and language-agnostic relations where events are entities (e.g., "Birth") that have two essential properties, a time and a location and where these properties have specific values of specific *types*[1]. Note also that we are assuming here that these canonical names are done in the process of KG construction, and thus a 'Birth' event, regardless how it was named, will in the end translate to the same event.

---

[1] While both 'human' and 'teacher' are concepts, a **human** is a type, while a teacher is not. In fact, a 'teacher' is (ontologically, or metaphysically an object of type human that we call (or label as) teacher when it is the agent of a teaching **activity**.

In our representation, therefore, everything is an entity and the relations come from a fixed set of primitive and linguistically agnostic set of relations (the set of primitive relations are shown in figure 2). How we come up with these relations is beyond the scope of this short paper but see Smith (2005) for a discussion.

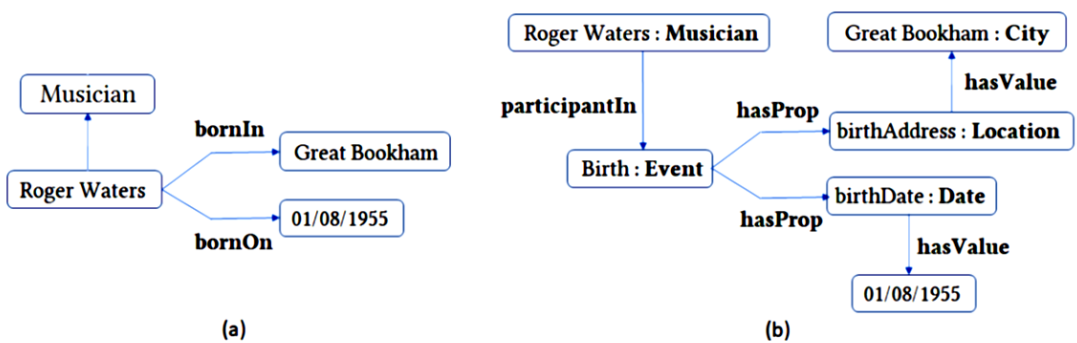

**Figure 1**: (a) A KG representing the facts expressed in "*The musician Roger Waters was born in Great Bokham on 01/08/1955*"; and (b) a language-agnostic KG representing the same facts.

Besides the primitive and linguistically agnostic representation, entities and attribute values in the knowledge graph of figure 1b are strongly-typed, where the types are assumed to exist in a strongly-typed hierarchy along the lines suggested in Saba (2020). Note that by making all entities typed we resolve the issue of separating knowledge graphs into two parts, one that has continuously updated information (⟨RogerWaters, LivesIn, London⟩) and one that has more static conceptual information such as ⟨RogerWaters, IsA, Musician⟩ (see Hao et. al., 2019 for a discussion on this issue).

| | |
|---|---|
| **eq**$(x, y)$ | $x$ is identical to $y$ |
| **isPartOf**$(x, y)$ | $x$ is part of $y$ |
| **inst**$(x, y)$ | $x$ instantiates $y$ |
| **hasProp**$(x, y)$ | $x$ inheres in $y$ |
| **exemp**$(x, y)$ | $x$ exemplifies $y$ |
| **dep**$(x, y)$ | $x$ depends on $y$ |
| **isA**$(x, y)$ | $x$ is a subtype of $y$ |
| **precedes**$(x, y)$ | $x$ precedes process $y$ |
| **participantIn**$(x, y)$ | $y$ participates in occurrent $x$ |
| **hasAgent**$(x, y)$ | $y$ is agent of occurrent $x$ |
| **hasObject**$(x, y)$ | $y$ is object of occurrent $x$ |
| **hasValue**$(x, y)$ | attribute $x$ has value $y$ |
| **realizes**$(x, y)$ | process $x$ realizes $y$ |

**Figure 2**: The set of primitive and linguistically agnostic relations that are used in the knowledge graph. These are the only relations used and all other abstractions are entities (e.g., events, properties, states, etc. all of which are reified/object-ified),

Moreover, entity alignment will now be more accurate since the embedding of [*RogerWaters*: **Musician**] will only match the same musician in another knowledge graph, even if the entity was labeled differently, e.g. [*GeorgeRogerWaters*: **Musician**]. Besides adding semantic constraints that will improve knowledge integration, types are language agnostic and thus, like primitive relations, are easy to translate across languages. In figure 3 we show the isomorphic Arabic and French equivalents of the KG in figure 1b above.

## 4 Evaluation

Aside from the simple alignment of knowledge graphs written in different languages or different domains, we show here how the ontologically grounded and linguistically agnostic representation helps in the problem of entity alignment. First, we construct embeddings for triples where a change is made in one of the entities or in the relation:

$e_1$ = EMBED($\langle$*RogerWaters, LivesIn, London*$\rangle$)
$e_2$ = EMBED($\langle$*RogerWaters, PlaceOfResidence, London*$\rangle$)
$e_3$ = EMBED($\langle$*RogerWaters, LivesIn, Chelsea*$\rangle$)
$e_4$ = EMBED($\langle$*RogerWaters, PlaceOfResidence, Chelsea*$\rangle$)

EMBED($\langle e_1, r, e_2 \rangle$) returns an embedding that is the sum of the vectors of $e_1$, $r$, and $e_2$. In table 1 below we show the cosine similarity **cosim**($e_i$, $e_j$) for i, j = 1,2,3,4 and for i ≠ j. The triples with a different entity (a different real-world fact) matched better than those with slightly different but semantically similar relation (i.e., same real-world fact).

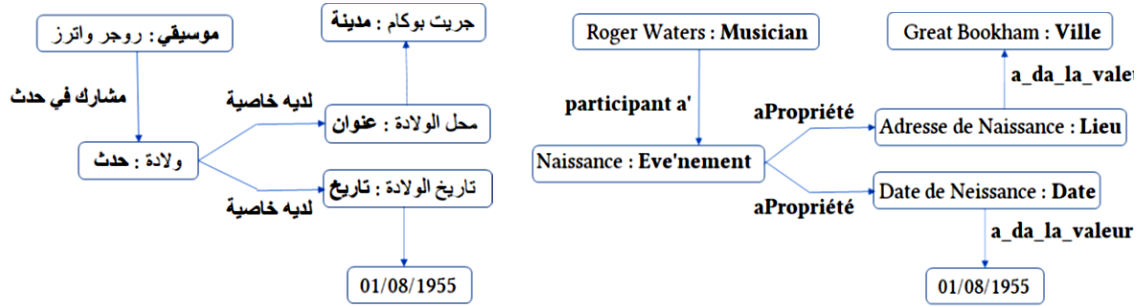

**Figure 3**: Since entity names, types, attribute values, and primitive relations are language agnostic, there's a straightforward automatic translation of the KG in figure 1b into isomorphic Arabic and French KGs.

Similar results were obtained by changing various semantically similar relations (e.g., **bornIn** vs. **placeOfBirth**, etc.)

The above shows that entity alignments across knowledge graphs would fail simply because of the ad-hoc labeling of relations in the knowledge graph. On the other hand, changing the location in the knowledge graph in 1b amounts to changing one embedding out of several that remain constant. In the example of figure 1b, a change in the location would result in a similarity of 0.688 only, and the alignment would clearly fail, as it should.

| COSINE_SIMILARITY(*emb1, emb2*) | 0.8853 |
| COSINE_SIMILARITY(*emb1, emb3*) | **0.9298** |
| COSINE_SIMILARITY(*emb1, emb4*) | 0.7989 |
| COSINE_SIMILARITY(*emb2, emb3*) | 0.8219 |
| COSINE_SIMILARITY(*emb2, emb4*) | **0.9204** |
| COSINE_SIMILARITY(*emb3, emb4*) | 0.8849 |

**Table 1**: Triples with different facts (locations) matched better than triples with the same facts (locations) but a relation that is worded slightly.

That is, an entity that is a participant in a birth event that happened in London should not match with an entity that is a participant in a birth event that happened in Chelsea, regardless of the entity name. Note that this true even in knowledge graphs in different languages (see figure 3), assuming, of course, that the embeddings of [London : **City**] and [لندن : مدينة] have a good cosine similarity, as one would expect.

## 5 Discussion

One important aspect to the representation we are suggesting is that it is language agnostic. This we claim is based on the fact that our representation has entities and primitive relations between them and that both of these are language agnostic. Thus the claim of universality is based two assumptions: (i) we are assuming that entities, including abstract entities such as those corresponding to properties, events, states, etc. are language-agnostic; (ii) we are assuming that our primitive relations (see figure 2) are also language agnostic. If both of these assumptions are correct, then our representation is language-agnostic, and the only remaining question would be "how universal are the primitive relations in figure 2?" A final answer to this question requires further experimentation.

Another important issue we could not discuss here for lack of space are the types that are associated with every entity and attribute value. These types are assumed to exist in a hierarchy of types that must also be language agnostic (that is, we are assuming that "the types of things we talk about/express facts about" are the same across languages). Admittedly, however, this claim might not be uncontroversial and further work needs to be done in this regard, although we believe the work of Saba (2020) is a step in the right direction. Another issue that should also be addressed is related to the mapping from natural language to our representation. As noted to us by one of anonymous reviewers, a fact such as "John sold the car to Bill" should, in theory, translate into the same sets of relations in the KG as the fact "Bill bought the car from John". While in both cases we will     have a language agnostic representation with reified abstract objects for the 'buying' and 'selling' events where Bill and John are participants, these two facts will only be equivalent if there were some meaning postulate that relates the 'selling' and 'buying' events.

## 5 Concluding Remarks

In this short paper we suggested an ontologically grounded and linguistically agnostic representation for knowledge graphs. This representation, we believe will solve the major challenges facing knowledge graphs today, namely the difficulty in continuous updating of factual information (which requires static conceptual information to be separated from the more dynamic information), and the difficulty of knowledge graph integration which requires very accurate entity and relation alignment. We argued that our representation offers a solution to these (essentially semantic) problems.

A final remark we would like to make is related to an excellent point made by one the anonymous reviewers, name that the representation and the method we propose will work if the construction of every KG follows our methodology. This is true, and so in essence the representation we are suggesting can be thought of as a new standard for a semantically rigorous knowledge graph methodology. Although this is part of future work, this will entail building a natural language interpreter that will ensure the translation of every KG into the canonical and language agnostic representation suggested in this paper.

## Acknowledgements

The feedback of colleagues at the Institute for Experiential AI as well as the suggestions of three anonymous reviewers are greatly appreciated.

## References

Nino B. Cocchiarella. 2001. Logic and Ontology, *Axiomathes*, 12: 117-150.

Donald Davidson. 1967. The Logical Form of Action Sentences, in N. Rescher (ed.) *The Logic of Decision and Action* (pp. 81-120) University of Pittsburgh Press.

Junheng Hao, Muhao Chen, et. al. 2019. Universal Representation Learning of Knowledge Bases by Jointly Embedding Instances and Ontological Concepts, In *25th ACM SIGKDD Conference on Knowledge Discovery and Data Mining (KDD '19)*

Richard K. Larson. 1998. Events and Modification in Nominals, *Semantics and Linguistic Theory*, Vol. 8, 145-168.

Yankai Lin, Zhiyuan Liu, Maosong Sun. 2016. Knowledge Representation Learning with Entities, Attributes and Relations, *In Proceedings of the Twenty-Fifth International Joint Conference on Artificial Intelligence* (IJCAI-16).

Friederike Moltmann. 2013. *Abstract Objects and the Semantics of Natural Language*, Oxford Uni Press.

Natasha Noy, Yuqing Gao, Anshu Jain, Anant Narayanan, Alan Patterson, and Jamie Taylor. 2019. Industry-Scale Knowledge Graphs: Lessons and Challenges, *Communications of The ACM* | August 2019, Vol. 62, No. 8, 36-43.

Walid Saba. 2020. Language and its commonsense: Where Formal Semantics Went Wrong, and Where it Can (and Should) Go, *Journal of Knowledge Structures and Systems* (JKSS), 1 (1):40-62

Barry Smith. 2005. Against Fantology, In Johann C. Marek & Maria E. Reicher (eds.), *Experience and Analysis*. Vienna: HPT: 153-170.

Yuxin Wang, Yuanning Cui, et. al. 2022. Facing Changes: Continual Entity Alignment for Growing Knowledge Graphs, In U. Sattler et al. (Eds.): *ISWC 2022, LNCS 13489*, pp. 196–213, 2022.

Rui Zhang, Bayu Distiawan Trisedya, Miao Li, Yong Jiang and Jianzhong Qi. 2022. A benchmark and comprehensive survey on knowledge graph entity alignment via representation learning, *The VLDB Journal*, 31: 143–1168.

Beibi Zhu, Tie Bao, Ridong Han, Hai Cui, Jiayu Han, Lu Liu and Tao Peng. 2023. An effective knowledge graph entity alignment model based on multiple information, *Neural Networks* 162:83–98

Renbo Zhu, Meng Ma and Ping Wang. 2021. RAGA: Relation-aware Graph Attention Networks for Global Entity Alignment, In *Proc. of Pacific-Asia Conference on Knowledge Discovery and Data Mining*.